# VLIPP: Towards Physically Plausible Video Generation with Vision and Language Informed Physical Prior

Xindi Yang[1]*, Baolu Li[2]*, Yiming Zhang[2], Zhenfei Yin[4,5]†, Lei Bai[3]†, Liqian Ma[6],
Zhiyong Wang[5], Jianfei Cai[1], Tien-Tsin Wong[1], Huchuan Lu[2], Xu Jia[2]†

[1]Monash University  [2]Dalian University of Technology  [3]Shanghai Artificial Intelligence Laboratory
[4]Oxford University  [5]The University of Sydney  [6]ZMO AI

## Abstract

*Video diffusion models (VDMs) have advanced significantly in recent years, enabling the generation of highly realistic videos and drawing the attention of the community in their potential as world simulators. However, despite their capabilities, VDMs often fail to produce physically plausible videos due to an inherent lack of understanding of physics, resulting in incorrect dynamics and event sequences. To address this limitation, we propose a novel two-stage image-to-video generation framework that explicitly incorporates physics with vision and language informed physical prior. In the first stage, we employ a Vision Language Model (VLM) as a coarse-grained motion planner, integrating chain-of-thought and physics-aware reasoning to predict a rough motion trajectories/changes that approximate real-world physical dynamics while ensuring the inter-frame consistency. In the second stage, we use the predicted motion trajectories/changes to guide the video generation of a VDM. As the predicted motion trajectories/changes are rough, noise is added during inference to provide freedom to the VDM in generating motion with more fine details. Extensive experimental results demonstrate that our framework can produce physically plausible motion, and comparative evaluations highlight the notable superiority of our approach over existing methods. More video results are available on our Project Page:* https://madaoer.github.io/projects/physically_plausible_video_generation/.

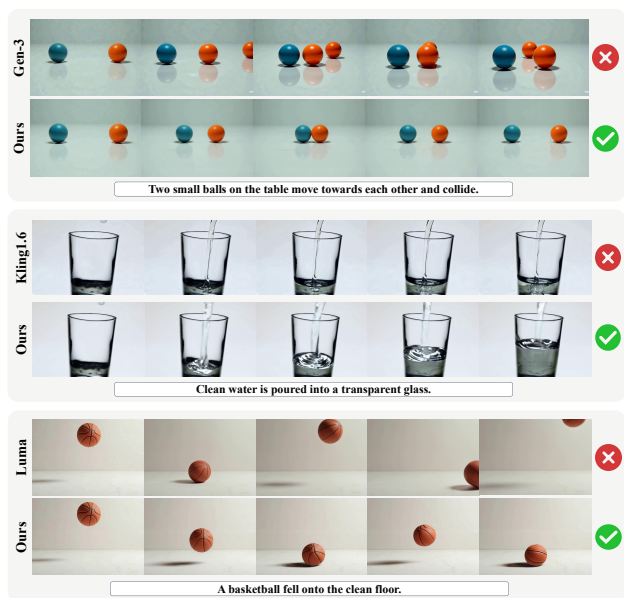

Figure 1. Existing commercial closed-source VDMs fail to generate physically plausible motion, whereas our video generation framework is able to achieve this by incorporating external physical prior knowledge.

## 1. Introduction

Video diffusion models (VDMs) trained on large-scale video datasets have made remarkable progress in terms of realism, demonstrating significant potential for various content creation applications. Despite the absence of explicit geometric modeling, the generated videos still exhibit coherent spatial relationships among objects, rich textured details, and realistic lighting effects, including reflections and shadows. Such qualities often make the generated videos nearly indistinguishable from real-world footages. This drives the research community to explore the potential of VDMs as world models. However, they still struggle with understanding the physical laws of the real world and generating videos that adhere to these principles.

Although existing VDMs can produce visually realistic videos, they fail to mimic the real-dynamic physical motions. As shown in Fig 1, even the current commercial closed-source VDMs struggle with the task of generating videos that conform to physical laws. PhyT2V [58] refines text prompts by incorporating detailed descriptions of phys-ical processes to guide VDMs in generating physically plausible videos. However, despite being pretrained on internet-scale real-world video–text pairs, VDMs do not inherently understand physical laws. This limitation arises from the gap and ambiguity between text descriptions and the actual motion in the video[33]. Moreover, VDMs tend to overfit the training data rather than developing a general understanding of physical laws[21]. Inspired by the success of graphics-based physical rendering, some methods have guided VDMs to generate physically plausible videos using simulations from graphics engines[18, 29, 32, 56, 64]; however, these approaches rely on the physical effects that graphics engines can simulate and incur high computational costs.

---

*Equal contribution  †Corresponding author

The gap and ambiguity between text and real-world motion makes it difficult to enable physically plausible video generation through detailed text descriptions alone. Moreover, it is challenging to gather scalable physical data for training due to the abstractness and diversity of physical phenomena. Consequently, a viable approach could be to model abstract physical laws as conditions for diffusion models. However, it is less practical to explicitly model the physics equation for every kind of motion. Instead, we resort to current large foundation models for their ability to "understand" basic physics[6] and reason about physical phenomena based on the knowledge they extract. For example, given two colliding balls, the Large Language Model (LLM) can *approximately* predict the paths of the balls after collision. Inspired by this observation, we propose a novel video generation framework that employs a Vision Language Model (VLM) to predict the path/change during a physics event, described by a given image and a text prompt.

In this paper, we propose VLIPP, a two-stage approach to incorporate physics as conditions into VDM, enabling the generation of physically plausible motion. In the first stage, the VLM serves as a *coarse-level motion planner*, while a VDM serves as a *fine-level motion synthesizer*. The idea of stage one is to utilize the chain-of-thought and the physics-aware reasoning of VLM planning to ensure that coarse-level motion trajectories approximately follow real-world physics dynamics. In stage two, we can generate fine-level motion using an image-to-video diffusion model conditioned by the approximated path/change planned by VLM from stage one. Note that the approximated path/changes are not in the level to tell the speed or acceleration of the motion. We choose an existing image-to-video model [7] to accept our coarse-level path/change, by injecting noise to the motion path during both the training and inference phases. Notably, during the VLM planning stage, generating entire physically plausible motion trajectories is not required. Instead, we leverage the generative priors of VDM to produce fine-level physically plausible videos based on coarse-level motion trajectories provided by the VLM. So that the detail-level motion such as speed, acceleration, and vibration are left to the VDM to synthesize.

We evaluate our physically plausible video generation framework with two major video physics benchmarks and achieved satisfactory results. Furthermore, we discuss and analyze multiple insightful design choices in our video generation framework, such as employing a motion planner tailored for different physics categories, and enhancing the robustness of diffusion model to noisy trajectories. Our contributions are summarized as follows:

1. We introduce a novel image-to-video generation framework for generating physically plausible videos by leveraging the VLM and VDM priors, significantly outperforming the contemporary competitors.
2. We propose a novel chain-of-thought and physics-aware reasoning approach in VLM, along with random noise injection in the latent space during video generation, which effectively improves both the generation quality and physical plausibility.
3. We conduct a comprehensive experiments and user studies to demonstrate the effectiveness and generalization of our framework in physically plausible videos generation.

## 2. Related Work

### 2.1. Physically Plausible Visual Content Generation

Generating physically plausible videos offers substantial value to real-world applications such as scientific simulations[43], robotics [4, 59], and autonomous driving [10, 48]. Traditional graphics pipelines rely on simulation systems to model physical phenomena [36, 42]. Inspired by these approaches, recent studies [18, 29] have performed dynamic simulations in image space based on physical engines. Furthermore, some methods [56, 64] incorporate physical priors into 3D representations to enable the synthesis of physically plausible motions. However, these rule-based or solver-based simulators face limitations in expressiveness, efficiency, generalizability, and parameter tuning. Furthermore, these simulators require significant expertise, rendering them inaccessible and unfriendly for users.

In addition, Some studies have explored VDMs for generating physically plausible videos. Li *et al*. [24] models natural oscillations and swaying in frequency-domain. A downstream rendering module then animates static images based on the generated motion information. PhysDiff [62] introduces physical simulator as constrain into the diffusion process by projecting denoised motion of a diffusion step into a physically plausible motion. These methods mainly focus only on specific types of physical motion and do not establish a generalizable approach for generating physically plausible videos.

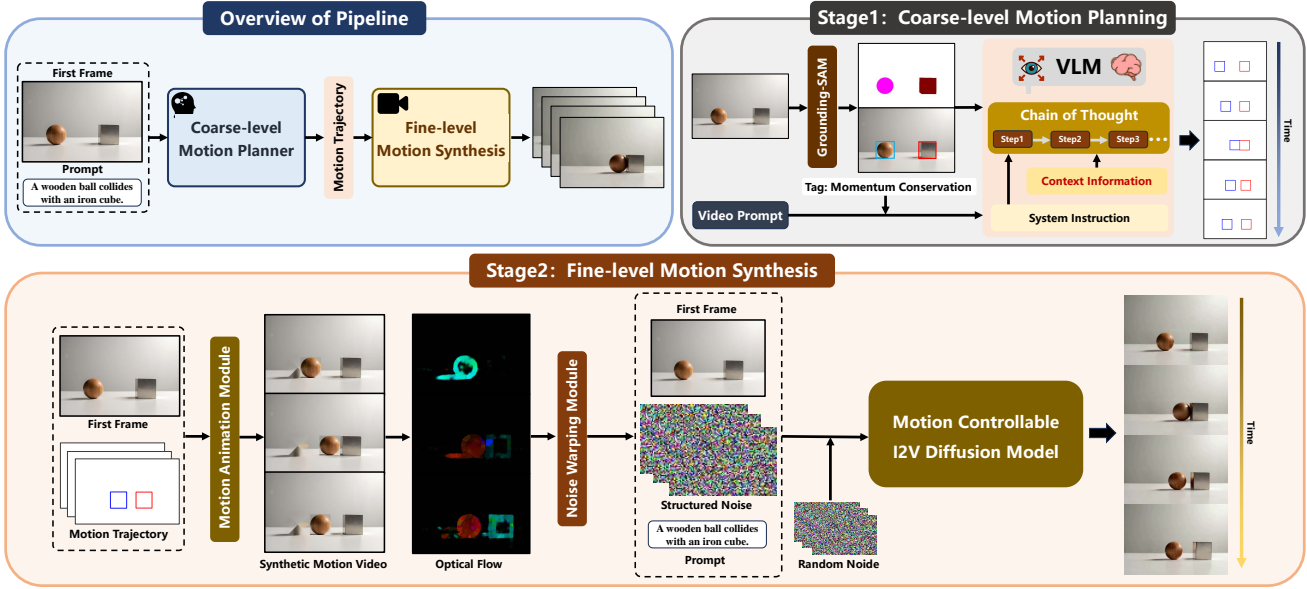

Figure 2. The illustration of our physically plausible image-to-video generation pipeline. Our pipeline consists of two stages. In the first stage, the VLM generates a coarse-grained, physically plausible motion trajectory based on the provided input conditions. In the second stage, We simulate a synthetic video using the predicted trajectory to provide the motion condition. We then extract the optical flow from this video and convert it into structured noise. These conditions are fed into a motion controllable image-to-video diffusion model, and ultimately generates a physically plausible video.

## 2.2. Motion Controllable Video Generation

Existing studies commonly provide one of the following three types of motion control: bounding box control [19, 23, 30, 46, 52], point trajectory control [34, 37, 40, 50, 54] and camera control [2, 15, 47, 61]. Bounding box control provides object motion guidance by generating a sequence of bounding boxes that track the object's position over time. Point-trajectory control offers motion cues through point-based trajectories, enabling drag-style manipulation. Camera motion control guides video generation using explicit 3D camera parameters, ensuring consistency and realistic viewpoint changes. However, these approaches prioritize motion control but often overlook physical plausibility. To address the limitation, we propose a novel framework for physically plausible video generation that incorporates physics as conditions into video diffusion models.

## 2.3. Generation based on VLMs Planning

VLMs have exhibited robust capabilities in visual understanding and planning [27, 39, 65]. Their strong performance in domains such as robot path planning and video understanding shows their ability in understanding the real physical world. Prior work has successfully leveraged LLMs to guide the layout of images or videos, yielding promising results [25, 53]. VideoDirectorGPT [26] leverages LLMs for fine-grained scene-by-scene planning, explicitly controlling spatial layout to generate temporally consistent long videos. Pandora [55] utilizes LLMs for real-time control through free-text action commands, achieving domain generality, video consistency, and controllability. However, these efforts have yet to address interactions with real-world physical phenomena, such as collision, fall, and melting.

Moreover, the absence of visual information can cause severe hallucination issues in language models for spatial planning tasks, leading to problems like overlapping object boundaries, disproportionate scaling, and incorrect planning [20, 57]. In this paper, we propose utilizing VLMs as coarse-level motion planners within the image space and incorporate physics-aware reasoning and Chain of Thought (CoT) [51] into the inference process.

## 3. Method

**Task Fomulation**. In this paper, our goal is to enable an image-to-video diffusion model to generate physically plausible videos. Since VDMs rely more on memory and case-based imitation and struggle to understand general physical rules[21], the key challenge is how to incorporate physical laws into the models. To achieve this, we need to identify a method to incorporate physical principles into the video diffusion framework. Given an image $I \in R^{H \times W \times C}$ ($H$ is height, $W$ is width and $C$ is the number of channels) and a text description $d$ of possible events based on image $I$, our framework should infer a physics-compliant guidance as the input condition and synthesize a video that adheres to both physical laws and real-world dynamics.

**Overall Pipeline**. Overall pipeline of VLIPP is illustrated

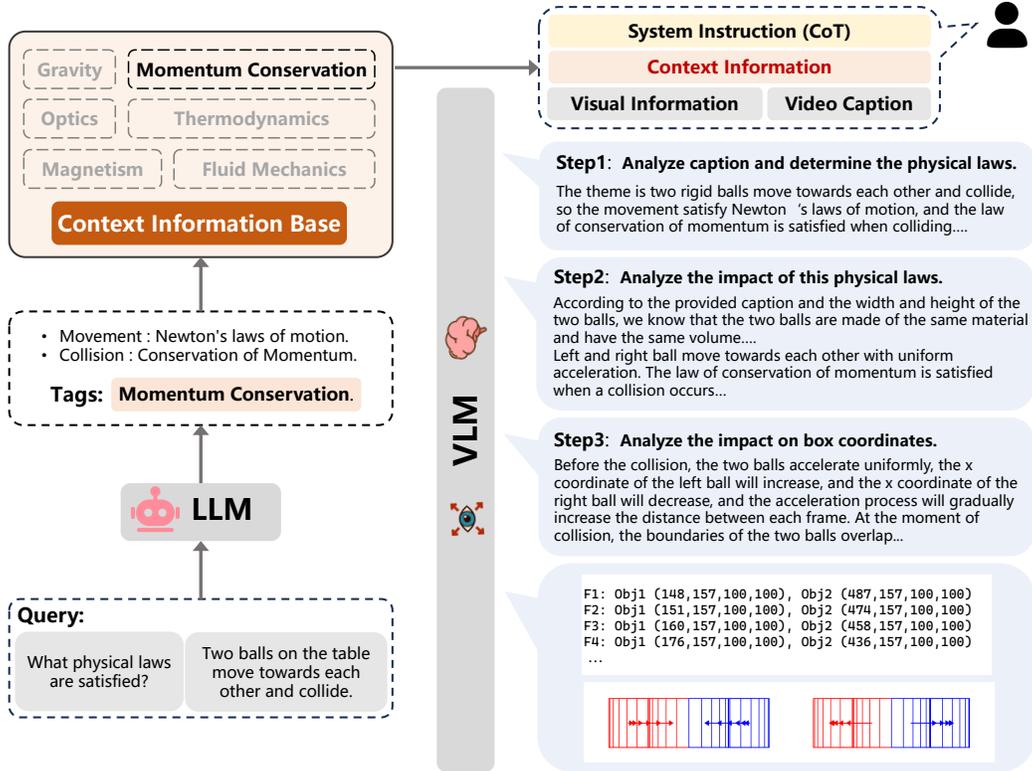

Figure 3. The illustration of chain-of-thought reasoning in the VLM for generating a coarse-grained motion trajectory. First off, the VLM determines the corresponding physical laws and its context for the given scene. Then, the VLM performs step-by-step reasoning to predict the physically plausible motions of objects in image space, leveraging physical context and chain-of-thought prompting. Finally, the VLM predicts bounding boxes according to real-world physics.

in Figure 2. In the first stage, the VLM conducts semantic analysis and physical attribute analysis on the given image $I$ and a description $d$ to obtain the bounding boxes of the objects in the scene, denoted as $b_1, b_2, ..., b_n$, along with the applicable physical laws $l$. Next, the VLM infers possible future physical scenarios in the current scene to derive the coarse-level motion trajectories of $p_1, p_2, ..., p_t$ in the image space. Finally, we utilize an image-to-video diffusion model to synthesize the detailed dynamics in the video.

### 3.1. VLM as a Coarse-Level Motion Planner

Our motivation is to incorporate physical laws as constraints into a video diffusion model to enhance the physical plausibility of the generated videos. To achieve this, we must identify a method to inject physical laws into the video diffusion model. Given a video description and the first frame, the task at this stage is to generate coarse-level motion trajectory aligned with physical laws.

**Scene Understanding**. In the real world, most physical phenomena arise from interactions among objects and their motion trajectories. We first initiate the process by identifying and locating objects within a scene. Inspired by recent studies [1, 9, 28] in VLMs for scene understanding, we employ GPT-4o [38] to recognize all objects that could be involved in physical phenomena as described in the text description $d$. These objects are subsequently detected and segmented using Grounded-SAM2 [41], yielding their bounding boxes. By leveraging the pretrained knowledge and common-sense reasoning capabilities of foundation models, we effectively determine the relevant objects in the scene.

**Physical-Aware Recognition**. To perform more effective reasoning in predicting the motion, it is necessary to determine what specific physical principle to apply in the given context. We utilize the pretrained prior of the LLM to determine the physical laws applicable to the current scene. Following the configuration in the physical benchmark [3, 31, 33], we currently classify common physical phenomena in videos into six categories: gravity, momentum conservation, optics, thermodynamics, magnetism, and fluid mechanics. Note that such list can be easily extended within our framework. Given a video description $d$, the LLM infers the physical law $l$ that governs the current scene. We provide the specific physical context information for VLM to enhance its understanding of physical laws [11]. Detailed context design is presented in the Appendix.

**Chain of Thought Reasoning in VLM**. Given the physical law $l$, an image $I$ and a video description $d$ for the scene, we prompt the VLM to predict the future bounding box positions of objects within the image-space. We choose to predict in the image space for two primary reasons. Firstly, motion in image space aligns more with our subsequent video synthesizer. Secondly, image space dynamics can effectively represent a wide range of real-world motions [29].

At a given time $t$, the predicted position of $i$-th object bounding box $b_i^t$ is denoted as $[x_t^i, y_t^i, w_t^i, h_t^i]$, where $(x_t^i, y_t^i)$ represents its top-left coordinate; $w_t^i$ & $h_t^i$ denote its width and height, respectively. Governed by the physical law, the four values of the bounding box may change over time. The VLM reasons the bounding box positions of $N$ future frames for every object $o_i$ based on the condition. To help VLM better understand physical laws, we adapt a chain-of-thought [51] into its reasoning, to significantly enhance its reasoning capabilities. As shown in Figure 3, we formulate our analysis of physical phenomena in videos as step-by-step reasoning: beginning with broad conceptual ideas and progressing to a detailed and practical examination:
1. Given the physical law $l$ and context information, the VLM analysis video caption and detail the physical law.
2. The VLM analyzes the potential interactions and movement of each object within the scene;
3. The VLM predicts the detailed changes in position and shape of the bounding box corresponding to each object over time.

Through the structured planning process, the VLM plans coarse-level motion trajectories for the objects, approximating real-world physics dynamics. In particular, our VLM infers the changes of object bounding boxes for next 12 frames, constrained by the token length limitation. To be compatible with the generation process of the chosen VDM in the next stage, these inferred 12 frames are further linearly interpolated to produce a total of 49 frames.

## 3.2. VDM Serves as a Fine-Level Motion Synthesizer

In the previous stage, the motion trajectory planned by the VLM is neither precise nor fully compliant with physical laws. On the other hand, while VDM may not be able to produce realistic global motion trajectories, it is able to generate sound motion in finer scale. In this stage, our key insight is that the VDM can refine the coarse-level motion to produce physically plausible motion that aligns with real-world dynamics with its powerful generative prior.

**Motion Animation**. To incorporate physical laws into the video diffusion model, we use the inferred coarse motion trajectory to guide the generation process of the diffusion model. Optical flow provides a unified representation of motion, and recent studies [7, 12, 13] have demonstrated its effectiveness in guiding diffusion models. Accordingly, we leverage the coarse-level motion trajectory to animate a synthetic motion sequence and derive the corresponding optical flow. Specifically, for each object $o_i$, we extract its bounding box from the first frame and move it to the bounding box location $b_i$ specified by the motion trajectory. To animate the change of shape (e.g., due to compression or expansion), we resize object $o_i$ according to the difference between $o_i$ and $o_{i+1}$ during inpainting. The synthetic motion video is generated as follows:

$$\hat{V}(t) = \text{Animation}(B, rs(o_0^0, b_t^0)...rs(o_0^i, b_t^i)) \quad (1)$$

where $\hat{V}(t)$ denotes the corresponding frame of inpainted video at timestep $t$, $B$ denotes the inpainted background with the foreground object removed, $o_i^0$ denotes the $i$-th object at timestep 0, $b_i^t$ represents the $i$-th bounding box at timestep $t$, and $rs$ denotes resize function.

**Structured Noise from Synthetic Video**. Optical flow is an effective representation for guiding VDMs [12, 13]. Follow prior work [7, 8], we employ RAFT[44] to extract optical flow from the synthetic video and formulate it as structural noise, which retains Gaussian properties. Given the synthetic video $\hat{V}(t) \in \mathbb{R}^{F \times C \times H \times W}$, we calculate its per-frame optical flow to get a structured noise tensor $Q \in \mathbb{R}^{F \times C \times H \times W}$. The structured noise enables the VDM to generate videos that exhibit motion patterns closely aligne with those in the optical flow, thereby improving the realism of the output.

**Noise Injection in Video Synthesis**. We adopt Go-with-the-Flow [7] as our video synthesis model, a fine-tuned CogVideoX [60], which is designed to accept structured noise $Q$ as input and synthesize videos that adhere to the implicit optical flow. The vanilla Go-with-the-Flow tends to tightly follow the provided structured noise $Q$. However, our $Q$ is derived from a coarse-level motion trajectory and may not be sufficiently accurate to follow the physical laws of the real world. To address this limitation, we inject noise during the inference phase to give more flexibility to the VDM to generate detail-level motion changes as

$$Q_i = \frac{(1-\gamma)Q_i + \zeta\gamma}{\sqrt{(1-\gamma)^2 + \gamma^2}} \quad (2)$$

where $Q_i$ is structured noise at $i$-th frames, $\zeta \in \mathbb{R}^{C \times H \times W}$ is Gaussian noise and $\gamma \in [0, 1]$. We set $\gamma = 0.4$ for even frame index and $\gamma = 0.6$ for odd frame index.

With this approach, the VDM is able to generate motion deviate from the coarse-level motion trajectory whenever necessary for producing high-quality fine-level motion.

## 4. Empirical Analysis and Disscusion

In this section, we conduct extensive experiments to demonstrate the effectiveness of our video generation framework compared to existing methods. We evaluate our approach on two established benchmarks for physically plausible video generation. Our framework consistently achieves superior performance across all benchmarks.

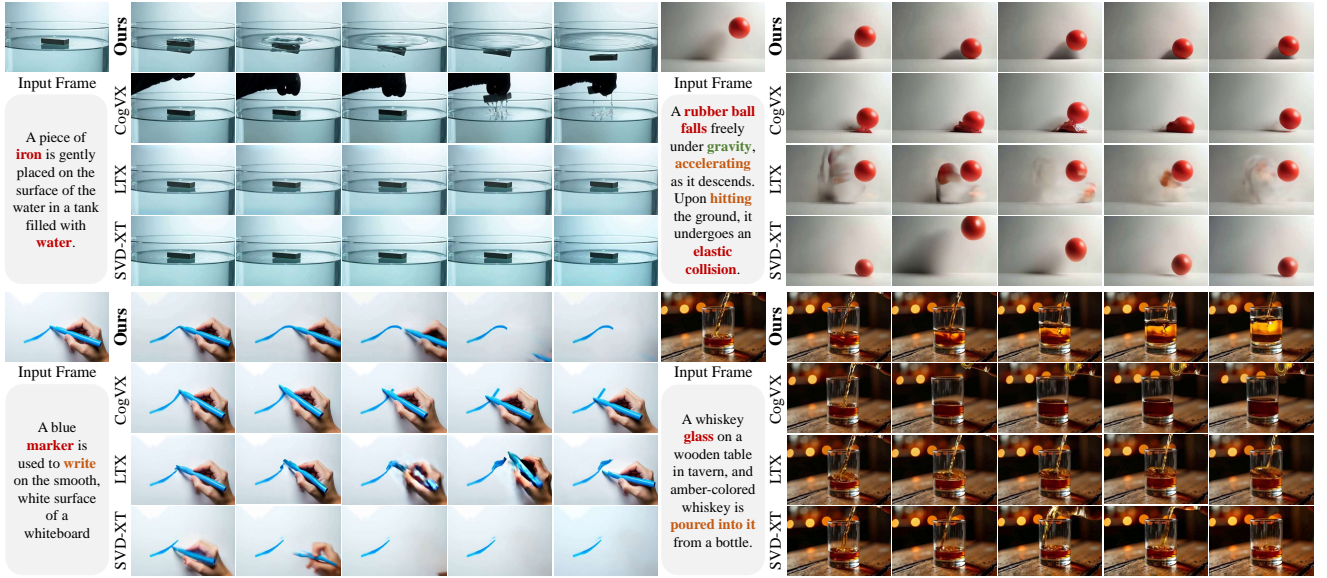

Figure 4. Visual comparisons of physically plausible video generation results from our framework, CogVideoX-I2V-5B [60], LTX-Video-I2V [14] and SVD-XT [5].

| Model | Mechanics(↑) | Optics(↑) | Thermal(↑) | Material(↑) | Average(↑) |
|---|---|---|---|---|---|
| CogvideoX-T2V-5B | 0.43 | 0.55 | 0.40 | 0.42 | 0.45 |
| LTX-Video-T2V | 0.35 | 0.45 | 0.36 | 0.38 | 0.39 |
| OpenSora | 0.43 | 0.50 | 0.44 | 0.37 | 0.44 |
| PhyT2V | 0.49 | 0.61 | 0.49 | 0.47 | 0.52 |
| LLM-Grounding Video Diffusion | 0.32 | 0.41 | 0.26 | 0.24 | 0.31 |
| CogvideoX-I2V-5B | 0.48 | 0.69 | 0.43 | 0.41 | 0.52 |
| SVD-XT | 0.46 | 0.68 | 0.48 | 0.41 | 0.52 |
| LTX-Video-I2V | 0.47 | 0.65 | 0.46 | 0.37 | 0.50 |
| SG-I2V | 0.52 | 0.69 | 0.51 | 0.39 | 0.54 |
| **Ours** | **0.55** | **0.71** | **0.60** | **0.53** | **0.60** |

Table 1. Quantitative results of VDMs on PhyGenBench.

### 4.1. Implementation Details

We propose a two-stage physically plausible image-to-video generation framework. In the first stage, we utilize ChatGPT-4o as the coarse-level motion planner. In the second stage, we utilize an open-source I2V model, Go-with-the-Flow [7], as a fine-level motion synthesizer. Unless otherwise specified, in all experiments, we generate each video with a resolution of $720 \times 480$ and 49 frames.

### 4.2. Benchmarks and Models

Traditional metrics in the visual domain, such as the Peak Signal-to-Noise Ratio (PSNR)[17], the Structural Similarity Index (SSIM)[49], the Learned Perceptual Image Patch Similarity (LPIPS)[63], the Fréchet Inception Distance (FID)[16] and the Fréchet Video Distance (FVD)[45], do not account for the physical realism of the generated videos [31, 33]. Recent studies have begun to address this limitation by developing benchmarks and metrics that evaluate physical realism. In this work, we adopt two benchmarks, described below.

**PhyGenBench** [31] categorizes physical properties into four domains: mechanics, optics, thermal, and material. It includes 27 physical phenomena, each governed by real world physical laws, reflected in 160 carefully designed text prompts. As PhyGenbench provides only text prompts, we adapt it to our image-to-video setting by generating a corresponding first frame for each prompt with FLUX[22]. We adhere to the predefined benchmark evaluation protocol, i.e., employing GPT-4o to assess the physical realism of the generated videos.

| Model | S.M.(↑) | F.D.(↑) | Optics(↑) | Magnetism(↑) | Thermodynamics(↑) | Average(↑) |
|---|---|---|---|---|---|---|
| Cogvideo-I2V-5B | 30.4 | 29.8 | 16.7 | 13.3 | 8.5 | 27.1 |
| SVD-XT | 21.9 | 20.5 | 6.8 | 8.4 | **17.1** | 19.1 |
| LTX-Video-I2V | 30.2 | 29.8 | 15.9 | 13.2 | 8.4 | 26.8 |
| SG-I2V | 34.6 | 31.2 | 15.9 | 13.1 | 8.4 | 29.7 |
| Ours | **42.3** | **34.1** | 16.9 | 13.4 | 8.8 | **34.6** |

Table 2. Quantitative results of physically plausible video generaion on Physics-IQ Benchmark. S.M. refers to Solid Mechanics, and F.D. refers to Fluid Dynamics.

**Physics-IQ** [33] comprises 396 real-world videos spanning 66 distinct physical scenarios. For each scenario, videos are recorded from three different perspectives and filmed twice under identical conditions to eliminate randomness. This benchmark evaluates real-world physical phenomena, including collisions, object continuity, occlusion, object permanence, and fluid dynamics. This benchmark assesses physical realism from semantic and temporal perspectives, using semantic metrics and visual metrics to compare generated videos against the real-world reference videos.

**Compared Models**. In the context of text-to-video generation, we compare our framework with CogVideoX-T2V-5B [60], LTX-Video-T2V [14], and OpenSora [66]. Moreover, we evaluate our framework against PhyT2V [58], which enhances physical realism by iteratively refining the prompt. For the image-to-video generation scenario, our framework is evaluated alongside CogVideoX-I2V-5B [60], SVD-XT [5], and LTX-Video-I2V [14]. Additionally, we conducted experiments in the motion-controllable setting. In this setting, we leverage the motion trajectory predicted by the VLM as a condition to guide VDM generation. We benchmark our approach against image-to-video motion controllable model, SG-I2V [35] and text-to-video motion controllable model LLM-grounded Video Diffusion Models [25]. The experimental details are presented in the Appendix.

### 4.3. Quantitative Evaluation

We begin with an empirical study on PhyGenBench and Physics-IQ, comparing our framework against widely adopted open-source models in the research community. Based on different physical properties, we categorize the benchmark samples accordingly. Additionally, we classify VDMs into text-to-video (T2V) diffusion models and image-to-video (I2V) diffusion models based on the input conditions.

In Table 1, we present our experimental results on PhyGenBench, evaluating different video generation models following its evaluation protocol. The results show that our framework achieves state-of-the-art performance across four different physical phenomena. **Our framework outperforms the best T2V method by an average of 15.3% and the best I2V method by 11.1%.** Specifically, our framework demonstrates significant advantages in the Mechanics, Thermal, and Material domains, outperforming the best I2V method by 5.7%, 17.6%, and 35.8%, respectively. These advantages are particularly evident in these three types of physical phenomena, which involve more substantial changes in motion, volume, or shape. Our framework is better equipped to understand and reason about bounding box sequences to represent these changes effectively.

Similarly, for the Physics-IQ benchmark, we evaluate the performance of different video generation models following its evaluation protocol. **Our framework achieves the best results across four different physical phenomena, with improvements of 22.2% in Solid Mechanics and 9.2% in Fluid Dynamics compared to the second-best models.** These significant improvements demonstrate the effectiveness of our framework in generating physically plausible videos.

### 4.4. Qualitative Evaluation

Figures 4 and 5 demonstrate a qualitative comparison between our video generation framework and baseline methods. Among all evaluated approaches, our framework consistently produces videos with the highest degree of physical realism. In the ball falling sample in Figure 4, while CogVideoX shows a bouncing effect, artifacts are present in the video; LTX-Video and SVD-XT exhibit motions that do not adhere to the laws of physics. In Figure 5, we analyze two examples from Physics-IQ. In the pouring water example, the baseline methods fail to show the simultaneous decrease in water level of the glass beverage dispenser and the increase in water level of the glass below; in the ball collision example, none of the baseline methods correctly depict the collision of balls. More videos are provided in the supplement.

### 4.5. Ablation Study

We perform an ablation study to evaluate the contributions of key components in our framework. We design four variants to analyze the effectiveness of different components in our framework.

1. **Ours w/o VLM Planner:** To assess the overall functionality of our framework, we replace the structured noise input of the VDM with random noise to evaluate the ef-fectiveness of the VLM planner.
2. **Ours w/o CI**: Keeping the overall structure unchanged, we remove the in-context information from the VLM.
3. **Ours w/o CoT**: Similarly, while keeping other components unchanged, we remove the CoT reasoning process from the VLM.
4. **Ours w/o CC**: Lastly, we remove both the in-context information and the CoT reasoning process from the VLM planner while maintaining all other components.

| Model | S.M.(↑) | F.D.(↑) | Optics(↑) | Magnetism(↑) | Thermodynamics(↑) | Average(↑) |
|---|---|---|---|---|---|---|
| Ours | **42.3** | **34.1** | **16.9** | **13.4** | **8.8** | **34.9** |
| w/o VLM Planner. | 16.3 | 20.8 | 13.4 | 5.8 | 5.6 | 16.2 |
| w/o C.I | 26.3 | 28.1 | 16.9 | 11.2 | 8.4 | 24.3 |
| w/o CoT | 21.4 | 26.9 | 16.1 | 8.6 | 6.9 | 21.0 |
| w/o C.C | 18.7 | 22.4 | 14.9 | 7.2 | 6.1 | 18.1 |

Table 3. Ablation study on VLM, in-context learning and COT. S.M. refers to Solid Mechanics, and F.D. refers to Fluid Dynamics.

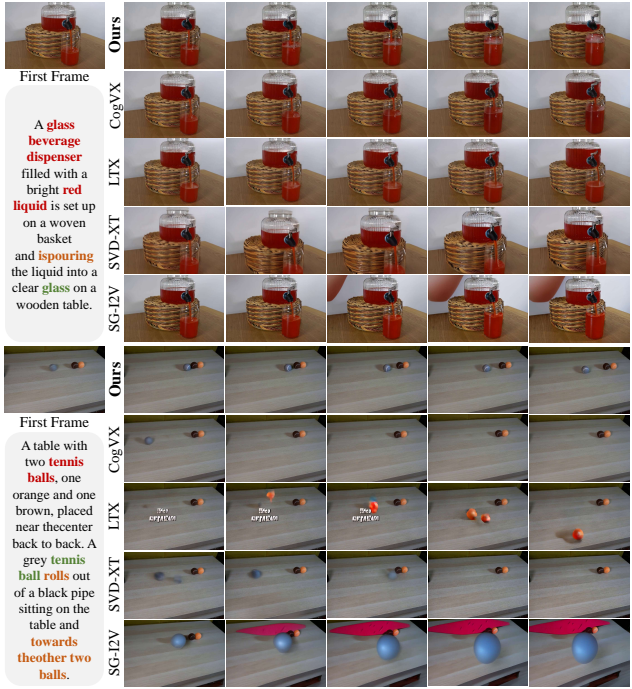

Figure 5. Visual comparisons of physically plausible video generation results from our framework, CogVideoX-I2V-5B, LTX-Video-I2V, SVD-XT and SG-I2V [35] in the Physics-IQ dataset.

Table 3 presents a quantitative comparison between our full method and these variants. Among all variants, **Ours w/o VLM Planner** shows the most significant performance drop, as removing the planner completely eliminates our ability in understanding the physical laws, leading to nearly random results. Notably, **Ours w/o CoT** exhibits a more pronounced decline compared to **Ours w/o CI**, indicating that the reasoning process in CoT enhances the understanding of physics. While in-context information contributes to the physical reasoning ability of VLM, compared to CoT it is less effective in preventing errors caused by VLM hallucination.

### 4.6. User Study

To complement our above evaluations, we conduct a user study to assess the subjective human perception of the generated videos. We follow the gold standard experimental approaches from psychophysics, a 2AFC paradigm, which means two-alternative-forced-choice [33]. In our case, participants completed a questionnaire in which they were presented with pairs of videos and asked to select the one that better aligned with their expectations of physical realism. Responses from 50 participants are summarized in Table 4. The result indicates a strong preference for videos generated by our framework over those from competitors. A detailed analysis of these findings follows in the subsequent discussion.

| Model | P.P.(↑) | V.R.(↑) |
|---|---|---|
| CogVideoX-I2V-5B | 34% | 40% |
| LTX-Video | 22% | 18% |
| Ours | **52%** | **48%** |

Table 4. User study statistics of the preference rate for Physical Plausibility (P.P.) & Visual Realism (V.R.).

### 4.7. Limitations

Although our framework can generate physically plausible videos, its performance remains constrained by the base model. Firstly, we cannot model physical events that cannot be represented by image space bounding box trajectories. For example, phenomena that involve intrinsic state changes of objects such as solid fragmentation and gas solidification. Moreover, our pipeline lacks 3D spatial perception. It is unable to understand the spatial relationships within the scene. Finally, the optical flow of small objects is prone to noise interference. This will cause our framework to generate ambiguous content. With the recent progress in video generation model, we anticipate that our framework will be further improved in generating videos under more challenging physical conditions.

# 5. Conclusion

Recently, VDMs have achieved great empirical success and are receiving considerable attention in computer vision and computer graphics. However, due to the lack of understanding of physical laws, VDMs are unable to generate physically plausible videos. In this paper, we introduce VLIPP, a novel two-stage physically plausible video generation framework that incorporates physical laws into video diffusion models through vision and language informed physical prior. Our experimental results demonstrate the effectiveness of our method compared to existing approaches.

# VLIPP: Towards Physically Plausible Video Generation with Vision and Language Informed Physical Prior

## Supplementary Material

## A. Coarse-Level Motion Planning Details

In this section, we present the experimental setting and details for reproducing the results. The main principle of our experimental setting is to fairly compare different Video Diffusion Models(VDMs) in generating physically plausible videos. Our adapt well-known open source model to serve as Compared Models. We disscuss these models in details.

1. **CogVideoX[60]**: CogVideoX is capable of performing both text-to-video generation and image-to-video generation. It provides two model variants, featuring 2 billion and 5 billion parameters, respectively. In our experiments, we configured CogVideoX to generate 49 frames with a resolution of 720×480.
2. **LTX-Video[14]**: LTX-Video is also capable of performing both text-to-video generation and image-to-video generation. In our experiments, we compared two versions of LTX-Video with corresponding methods. It can generate videos with 49 frames with a resolution of 768×512.
3. **SVD-XT[5]**: SVD-XT is capable of performing image-to-video generation. In our experiments, we configured SVD-XT to generate 25 frames with a resolution of 1024×576.
4. **SG-I2V[35]**: SG-I2V is a motion trajectory-guided image-to-video generation model. It is capable of generating bounding box-controllable videos with 14 frames with a resolution of 1024×576.
5. **LLM-grounded Video Diffusion Models[25]**: LLM-grounded Video Diffusion Models are capable of predicting future frame bounding boxes based on input prompts and injecting the box information in a training-free manner. In our experiments, we configured LLM-grounded to generate 24 frames with a resolution of 576×320.

We additionally present the Reasoning Template utilized during the stage 1 Coarse-Level Motion Planning process, as shown in Fig 6 and Fig 7. This includes system instructions to ensure the proper functioning of the chain of thought and provides the VLM with context information to guarantee the accuracy of predictions.

## B. Experiment Details

In this section, we present the experimental details of our benchmark, PhyGenBench[31] and Physics-IQ[33].

PhyGenBench comprises 160 prompts, spanning four domains of physical knowledge: Mechanics (40), Optics (50), Thermal (40), and Material (20), along with 27 types of physical laws. It also includes 165 objects and 42 actions. The evaluation focuses on two aspects: semantic alignment and physical commonsense alignment. The degree of semantic alignment is assessed by extracting objects and actions from the prompts using a Vision-Language Model (VLM), determining whether the objects appear, and evaluating based on the presence of objects and the occurrence of actions. The degree of physical commonsense alignment is determined through a three-step process: detecting whether the physical phenomena occur and whether the order of occurrence is correct; and finally conducting an overall naturalness evaluation.

Physic-IQ categorizes real-world physical laws into Solid Mechanics, Fluid Dynamics, Optics, Magnetism, and Thermodynamics, encompassing 114, 45, 24, 6, and 9 videos, respectively. The evaluation approach is twofold, focusing on physical comprehension and visual authenticity. Physical comprehension is determined by identifying the timing, location, and frequency of actions, ultimately calculating the mean squared error between corresponding pixels in the generated and real frames to derive a physical comprehension score. Visual authenticity is evaluated using a Vision-Language Model (VLM), employing the gold standard experimental method from psychophysics. The VLM receives pairs of real and generated videos of the same scene in random order and is tasked with identifying the real scene, a design intended to reflect visual authenticity.

During the experimental phase of this paper, we utilized the prompts provided by the PhyGenBench dataset to infer the initial frame's prompts using an LLM, which were then generated by FLUX[22]. To ensure fairness in comparison, all I2V models were supplied with the same initial frame image. Given that different models produce videos with varying numbers of frames, a uniform sampling ratio was applied during the testing phase to extract key frames consistently across all models.

## C. More Qualitative Results

In this section, we further demonstrate examples of the proposed framework across various scenarios using prompts from the PhyGenBench dataset. Fig 8, 9, and 10 show video generation results driven by physical conditions like fluid dynamics and thermodynamics, while Fig 11 highlights effects in light reflection scenarios.

# Physics-Aware Reasoning Template (Gravity/Moumentum)

## System Instruction

You are an expert in real-world physical motion. Your task is to predict video bounding boxes. You will receive a reference image, its segmentation map, the initial bounding box where motion starts, and a textual description of the video content. Your task is to predict the bounding box coordinates for the next 13 frames. When predicting, ensure that different bounding boxes do not overlap or exceed the frame boundaries.

Each frame should be represented as: `[{'id': unique object identifier, 'name': object name, 'box': [box top-left x-coordinate, box top-left y-coordinate, box width, box height]}, ...]`.Your predicted frames must cover the entire described scene. Critical moments in the movement must be included in the predicted sequence, such as the moment of collision.You need to consider the material and properties of the moving objects in the video. You only need to focus on the objects provided to you in the first frame, and no need to pay attention to anything else. Your predictions should infer which physical laws apply, such as Gravity, Conservation of Momentum, Fluid Mechanics, Thermodynamics, Magnetic Force, etc. For example, under the action of gravity, the ball will accelerate and rebound after colliding with the ground. When two balls move towards each other, the material, volume, density, etc. of each ball must be considered, and the collision process satisfies the law of conservation of momentum. Assume objects move and interact based on real-world physics. The camera follows perspective geometry. Even if an object disappears and reappears, its bounding box should maintain the same ID across frames. If necessary, you may make reasonable assumptions.Before generating bounding box coordinates for each frame, provide step-by-step reasoning.

## Context Information

**User:** Caption: Generate a high-quality video of a wooden ball dropping from a height of about 1 meter onto a smooth, solid table.Capture the physics of the ball's descent, including its acceleration due to gravity, the moment of impact with the table, and the subsequent bounce. Show the ball compressing slightly upon impact, then rebounding, losing a bit of height with each bounce until it comes to rest. Include subtle details like the sound of the impact, the slight vibration of the table, and the ball's rotation as it moves.
initial_boxes: [{'id': 0, 'name': 'ball', 'box': [303.0, 82.2, 108, 108]}]

**Assistant:** Reasoning:
  **Step1-Analysis Caption and determine the physical laws to be satisfied**: As can be seen from the caption, the topic of this study is about the free fall of a small ball, so this topic needs to satisfy Newton's laws of motion, focusing on the impact of gravity.
  **Step2-Analyze the impact of this physical law**: In accordance with Newton's laws of motion, gravity will cause the small ball to perform free fall motion, vertically accelerate and hit the table, and rebound after the collision. Due to energy loss, the height of each rebound will decrease, which is in line with the law of conservation of energy and momentum.
  **Step3-Analyze the impact on box coordinates**: The horizontal motion of the ball is negligible, so the x coordinate remains basically unchanged. Due to the vertical acceleration, the y coordinate of the small ball will gradually increase and the interpolation of the y coordinate between each two frames will also gradually increase. After colliding with the table, the small ball rebounds, the y coordinate decreases, and the small ball decelerates vertically upward. Subsequently, the small ball continues to accelerate vertically and collides with the table for the second time. After the third collision, the small ball is stationary on the table, and the y coordinate remains unchanged. Since the ball is a rigid body, it will be slightly compressed at the moment of collision with the desktop. Therefore, the width in the box coordinates remains unchanged, the height decreases slightly, and then quickly returns to normal. The width and height will not change for the rest of the time.
  **Predict the coordinates for the next 13 frames:**
  Frame 1: [{'id': 0, 'name': 'wooden ball', 'box': [303, 82, 108, 108]}]
  Frame 2: [{'id': 0, 'name': 'wooden ball', 'box': [303, 137, 108, 108]}]
  Frame 3: [{'id': 0, 'name': 'wooden ball', 'box': [303, 184, 108, 108]}]
  Frame 4: [{'id': 0, 'name': 'wooden ball', 'box': [303, 313, 108, 108]}]
  Frame 5: [{'id': 0, 'name': 'wooden ball', 'box': [303, 255, 108, 108]}]
  Frame 6: [{'id': 0, 'name': 'wooden ball', 'box': [303, 284, 108, 104]}]
  Frame 7: [{'id': 0, 'name': 'wooden ball', 'box': [303, 313, 108, 108]}]
  Frame 8: [{'id': 0, 'name': 'wooden ball', 'box': [303, 298, 108, 108]}]
  Frame 9: [{'id': 0, 'name': 'wooden ball', 'box': [303, 305, 108, 108]}]
  Frame 10: [{'id': 0, 'name': 'wooden ball', 'box': [303, 313, 108, 104]}]
  Frame 11: [{'id': 0, 'name': 'wooden ball', 'box': [303, 313, 108, 108]}]
  Frame 12: [{'id': 0, 'name': 'wooden ball', 'box': [303, 313, 108, 108]}]
  Frame 13: [{'id': 0, 'name': 'wooden ball', 'box': [303, 313, 108, 104]}]

Figure 6. Physics-Aware Reasoning Template for Rigid Body Motion

**Physics-Aware Recognition Template** (Fluid)

### System Instruction

You are an expert in real-world physical motion. Your task is to predict video bounding boxes. You will receive a reference image, its segmentation map, the initial bounding box where motion starts, and a textual description of the video content. Your task is to predict the bounding box coordinates for the next 13 frames. When predicting, ensure that different bounding boxes do not overlap or exceed the frame boundaries.

Each frame should be represented as: `[{'id': unique object identifier, 'name': object name, 'box': [box top-left x-coordinate, box top-left y-coordinate, box width, box height]}, ...]`. Your predicted frames must cover the entire described scene. Critical moments in the movement must be included in the predicted sequence, such as the moment of collision. You need to consider the material and properties of the moving objects in the video. Your predictions should infer which physical laws apply, such as Gravity, Conservation of Momentum, Fluid Mechanics, Thermodynamics, Magnetic Force, etc. For example, under the influence of fluid mechanics, when pouring water into a cup, the water level in the cup should rise and ripples should occur. The top boundary of a liquid should move when it's being poured, while the bottom edge of the liquid (for instance, in a container) should stay fixed. When an object (such as a liquid) is in a container and is being poured, the bottom boundary of the liquid typically stays fixed. Assume objects move and interact based on real-world physics. The camera follows perspective geometry. Even if an object disappears and reappears, its bounding box should maintain the same ID across frames. If necessary, you may make reasonable assumptions. Before generating bounding box coordinates for each frame, provide step-by-step reasoning.

### Context Information

**User:** Caption: The video shows a close-up of a clear glass being filled with tea. The tea is being poured from above, and we can see the stream of tea hitting the bottom of the glass and causing ripples and splashes. The background is a plain white surface, which contrasts with the transparency of the glass and the clarity of the tea. The glass is cylindrical in shape and appears to be of a standard size for a drinking glass.

initial_boxes: [{'id': 0, 'name': 'tea', 'box': [241.8, 335.3, 165, 88]}]

**Assistant:** Reasoning:

**Step1-Analysis Caption and determine the physical laws to be satisfied**: As can be seen from the caption, the theme of this session is to pour tea into a glass teacup. Therefore, the tea is affected by gravity during its fall, and the state of the tea in the cup satisfies fluid mechanics.

**Step2-Analyze the impact of this physical law**: Tea falls into the cup under the influence of gravity. According to fluid mechanics, the water level in the cup will gradually rise, and the water surface will ripple due to the pouring of water.

**Step3-Analyze the impact on box coordinates**: When tea is poured into the cup, under the restriction of the cup, the x coordinate of the tea box will not change much. As the water level rises, the y coordinate will gradually decrease. As the water level rises, the overall volume of the tea gradually increases, the height of the box will also increase, and the width will change with the boundary of the cup.

**Predict the coordinates for the next 13 frames:**

Frame 1: [{'id': 0, 'name': 'tea', 'box': [241.8, 335.3, 165, 88]}]
Frame 2: [{'id': 0, 'name': 'tea', 'box': [241.2, 330.0, 165, 94]}]
Frame 3: [{'id': 0, 'name': 'tea', 'box': [241.0, 322.2, 165, 102]}]
Frame 4: [{'id': 0, 'name': 'tea', 'box': [241.3, 317.6, 165, 105]}]
Frame 5: [{'id': 0, 'name': 'tea', 'box': [240.6, 313.7, 166, 111]}]
Frame 6: [{'id': 0, 'name': 'tea', 'box': [239.1, 305.6, 167, 118]}]
Frame 7: [{'id': 0, 'name': 'tea', 'box': [237.7, 301.1, 170, 123]}]
Frame 8: [{'id': 0, 'name': 'tea', 'box': [236.8, 295.3, 172, 129]}]
Frame 9: [{'id': 0, 'name': 'tea', 'box': [235.7, 290.2, 172, 134]}]
Frame 10: [{'id': 0, 'name': 'tea', 'box': [235.1, 283.3, 173, 140]}]
Frame 11: [{'id': 0, 'name': 'tea', 'box': [234.0, 278.3, 175, 145]}]
Frame 12: [{'id': 0, 'name': 'tea', 'box': [233.1, 272.0, 177, 152]}]
Frame 13: [{'id': 0, 'name': 'tea', 'box': [231.7, 268.0, 179, 156]}]

Figure 7. Physics-Aware Reasoning Template for fluid dynamics and thermodynamics.

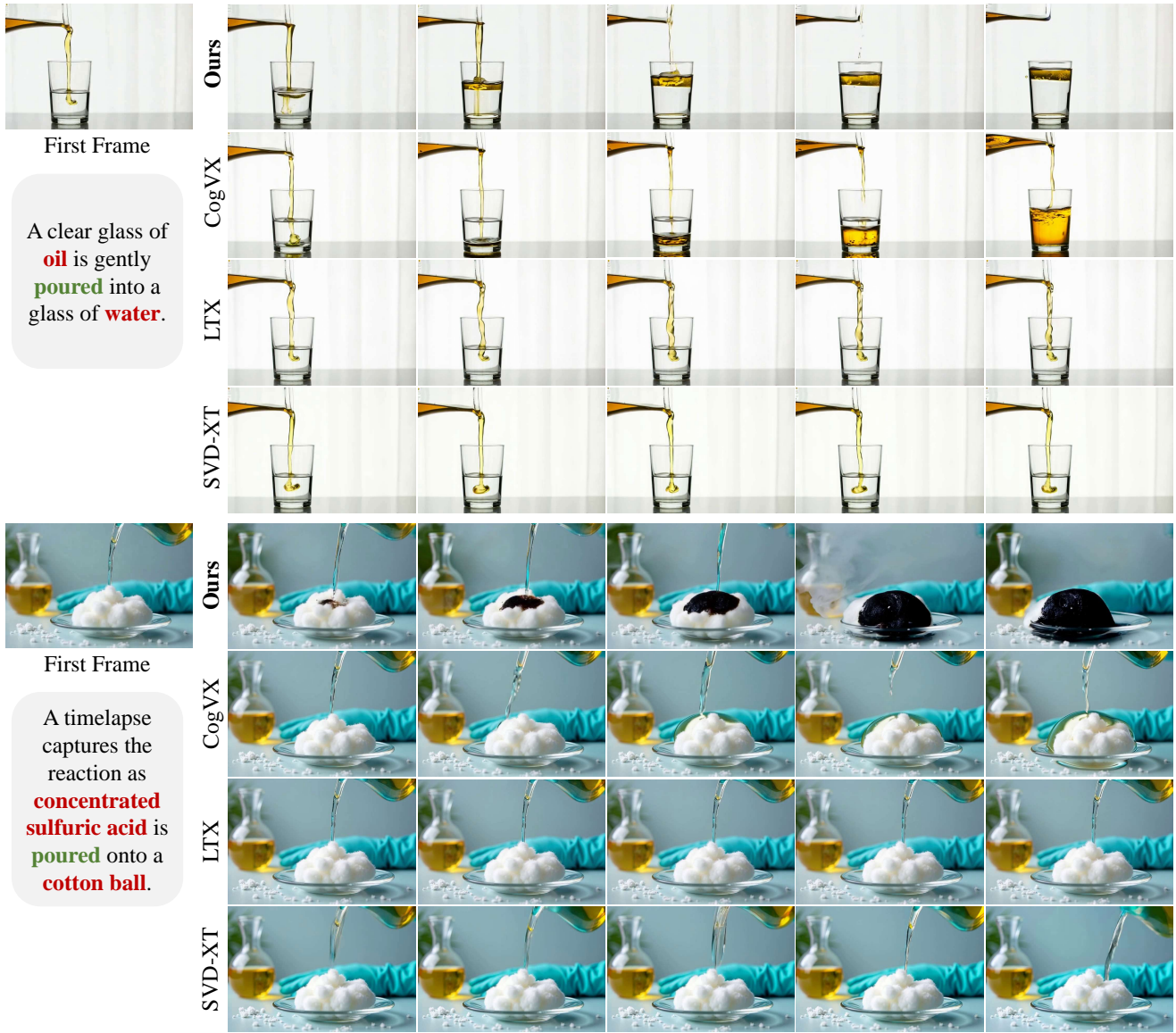

Figure 8. More examples of generated videos related to fluid dynamics and thermodynamics.

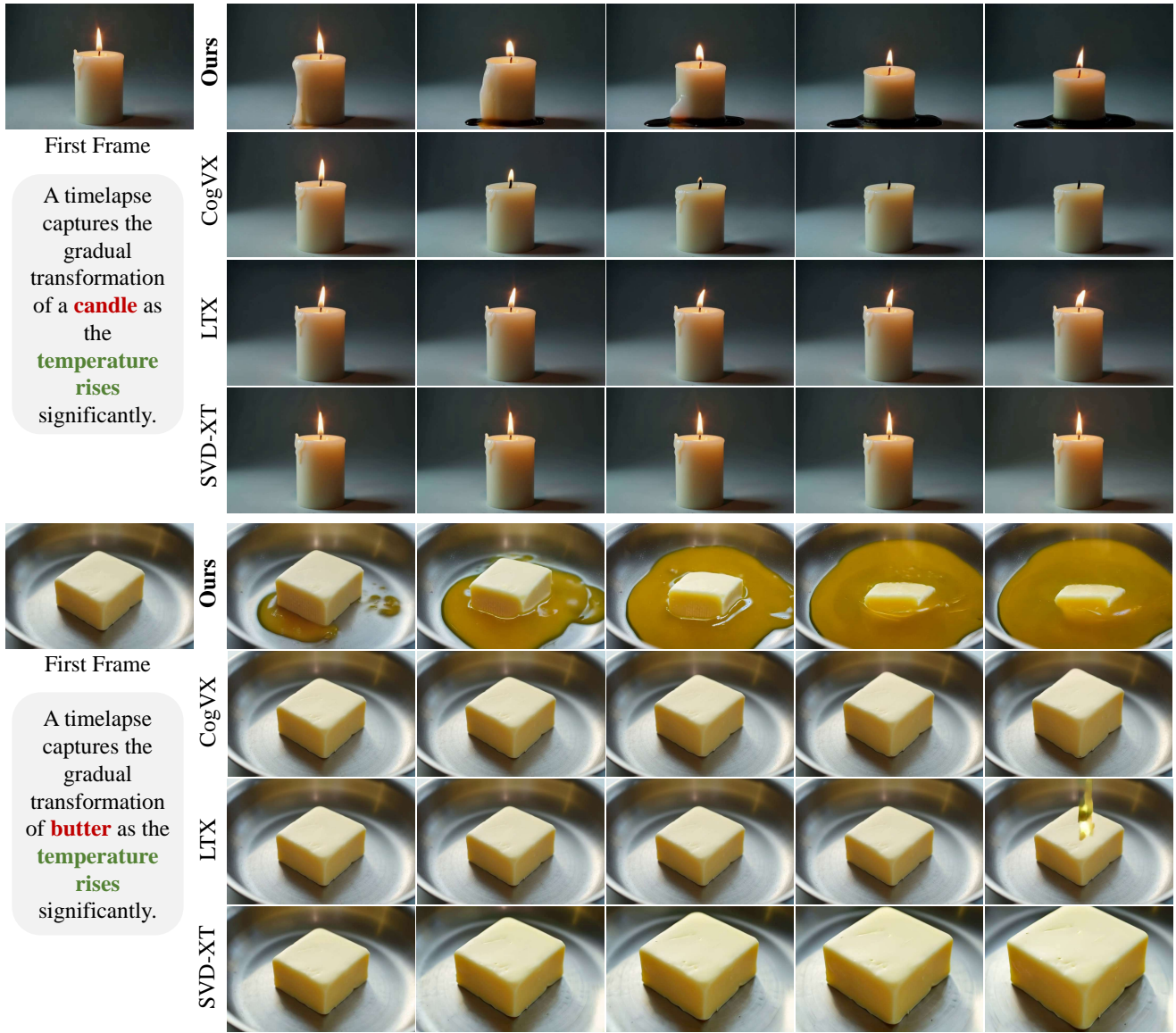

Figure 9. More examples of generated videos related to thermodynamics.

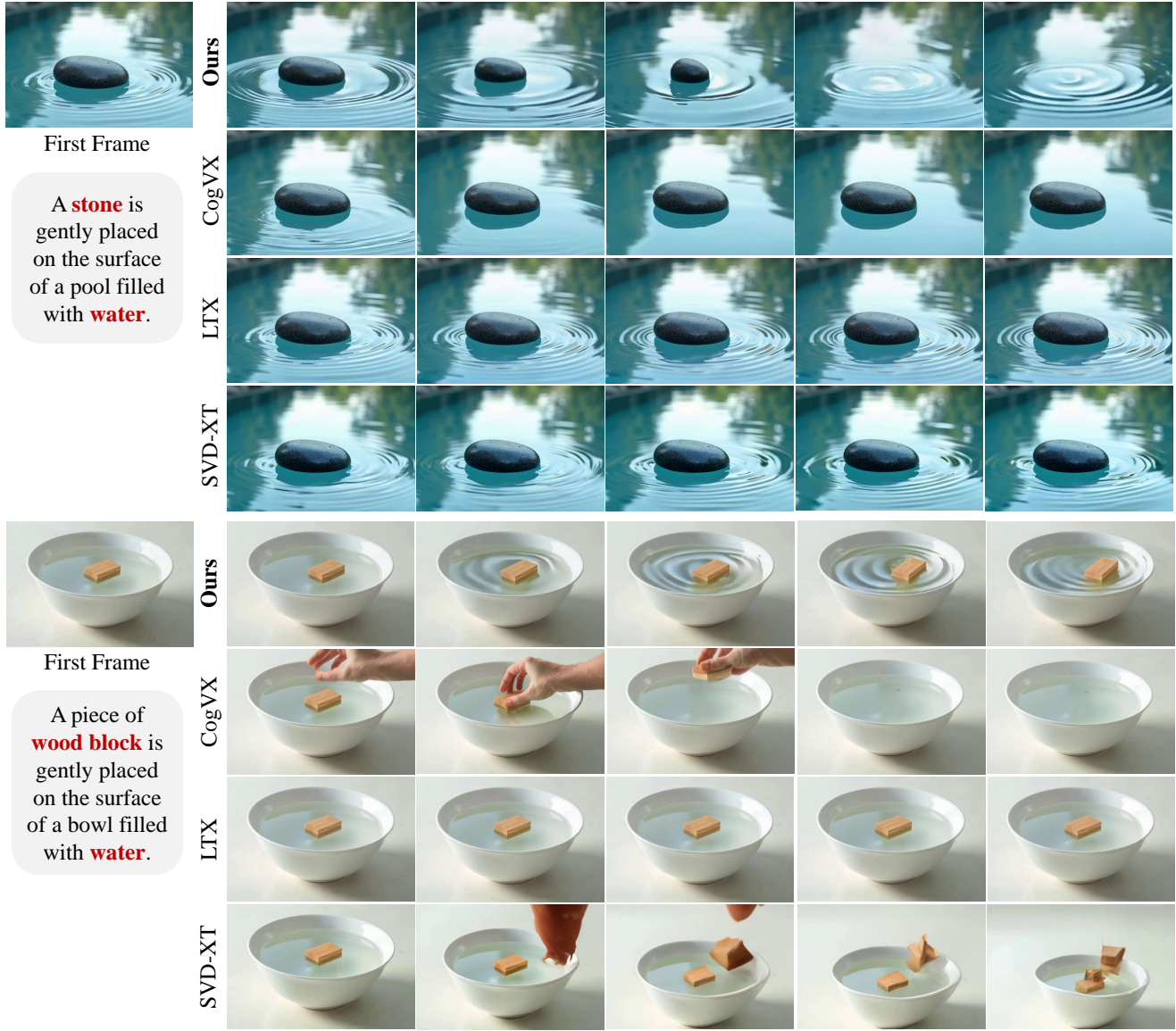

Figure 10. More examples of generated videos related to fluid dynamics.

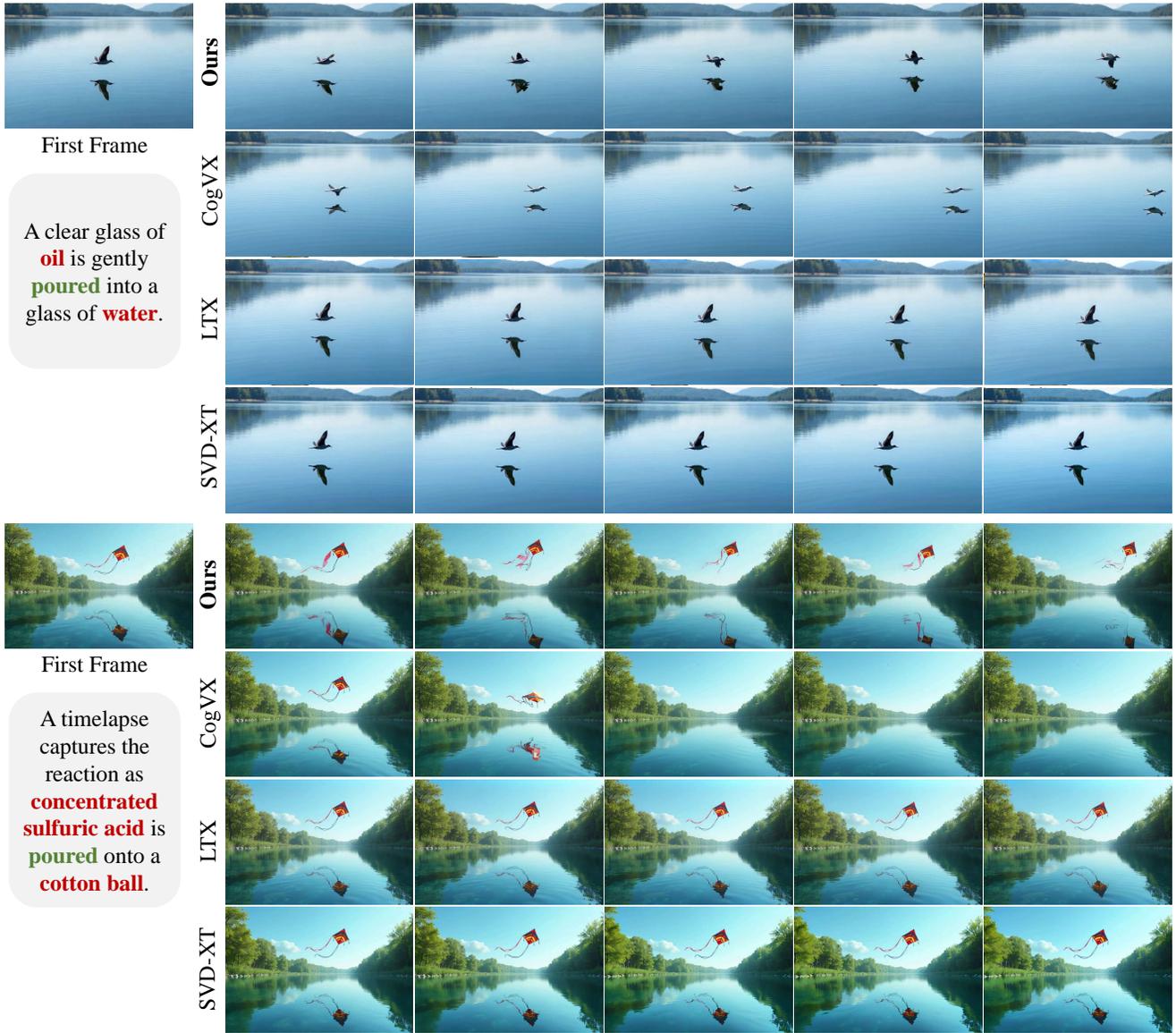

Figure 11. More examples of generated videos related to optics.